\documentclass[letterpaper, 10 pt, conference]{ieeeconf}  

\IEEEoverridecommandlockouts                              

\usepackage{graphicx}
\usepackage{amsmath}
\usepackage{amssymb}
\usepackage{booktabs}
\usepackage{multirow}
\usepackage{xcolor}
\usepackage{url}
\usepackage{cite}

\usepackage{algorithm}
\usepackage{algpseudocode}

\newcommand{\ours}{ArborSplat}

\title{\LARGE \bf
ArborSplat: Online Semantic Gaussian Splatting SLAM\\
for Orchards
}

\author{%
Alessandro Masini$^{1}$,
Matteo Frosi$^{*1}$,
Mirko Usuelli$^{1}$,
and Matteo Matteucci$^{1}$%
\thanks{$^{*}$ Corresponding author}
\thanks{%
$^{1}$All authors are with the Dipartimento di Elettronica,
Informazione e Bioingegneria (DEIB), Politecnico di Milano,
20133 Milano, Italy.
Email:
{\ttfamily\small alessandro.masini@mail.polimi.it};
{\ttfamily\small
\{name.surname\}@polimi.it}.%
}%
}

\usepackage{tikz}
\newcommand\submittedtext{%
  This work has been submitted to the IEEE for possible publication. Copyright may be transferred without notice, after which this version may no longer be accessible.}
\newcommand\submittednotice{%
\begin{tikzpicture}[remember picture,overlay]
\node[anchor=south,yshift=15pt] at (current page.south) {\fbox{\parbox{\dimexpr0.95\textwidth-\fboxsep-\fboxrule\relax}{\submittedtext}}};
\end{tikzpicture}%
}

\begin{document}

\bstctlcite{IEEEexample:BSTcontrol}

\maketitle
\thispagestyle{empty}
\pagestyle{plain}
\submittednotice

\begin{abstract}

Orchard robots need maps that preserve small but semantically important structures such as trunks, trellises, and fruit. 3D Gaussian Splatting (3DGS) SLAM achieves high photometric fidelity. However, its optimization remains appearance-driven, and transferring image semantics to 3D points is unreliable for thin structures, whose pixels may receive depth from background surfaces. We present \ours{}, an online semantic 3DGS SLAM system that tracks with LiDAR odometry and optimizes semantics directly on the Gaussian map, constrained by class-specific height bands above a ground plane fitted to each keyframe's stereo point cloud, and fuses multi-view evidence into a semantic point cloud online while rejecting labels inconsistent with the local ground surface or with monocular depth. Class-constrained refinement reserves Gaussian capacity for underrepresented structures and, under reduced budgets, increases training-view accuracy on tree classes. We evaluate the approach on apple and pear orchards during dormancy, flowering, and harvesting.
On full routes, it keeps ATE below 0.5\,m on all 12 traversals. On shared 301-frame segments, it exceeds SGS-SLAM and GS3LAM by 0.23 to 0.50 training-view and 0.15 to 0.36 held-out mIoU while running 1.7 to 7.5 times faster, whereas SemGauss-SLAM runs out of GPU memory on all six.

\end{abstract}

\section{INTRODUCTION}

Labor shortage is becoming a limiting factor for agricultural production~\cite{kootstra2021selective}, pushing orchards toward robotic automation throughout the year. Orchard tasks depend on structures that occupy a small fraction of the image: pruning in dormancy requires distinguishing woody structure from support wire, spraying requires separating canopy from trellis, and harvesting requires localizing fruit within both. Robots thus need maps that are semantic as well as spatial, that hold in every season, and whose quality is judged on these small structures rather than on dominant scene regions.

\begin{figure*}[!t]
    \vspace{.1em}
   \centering
   \includegraphics[width=\textwidth]{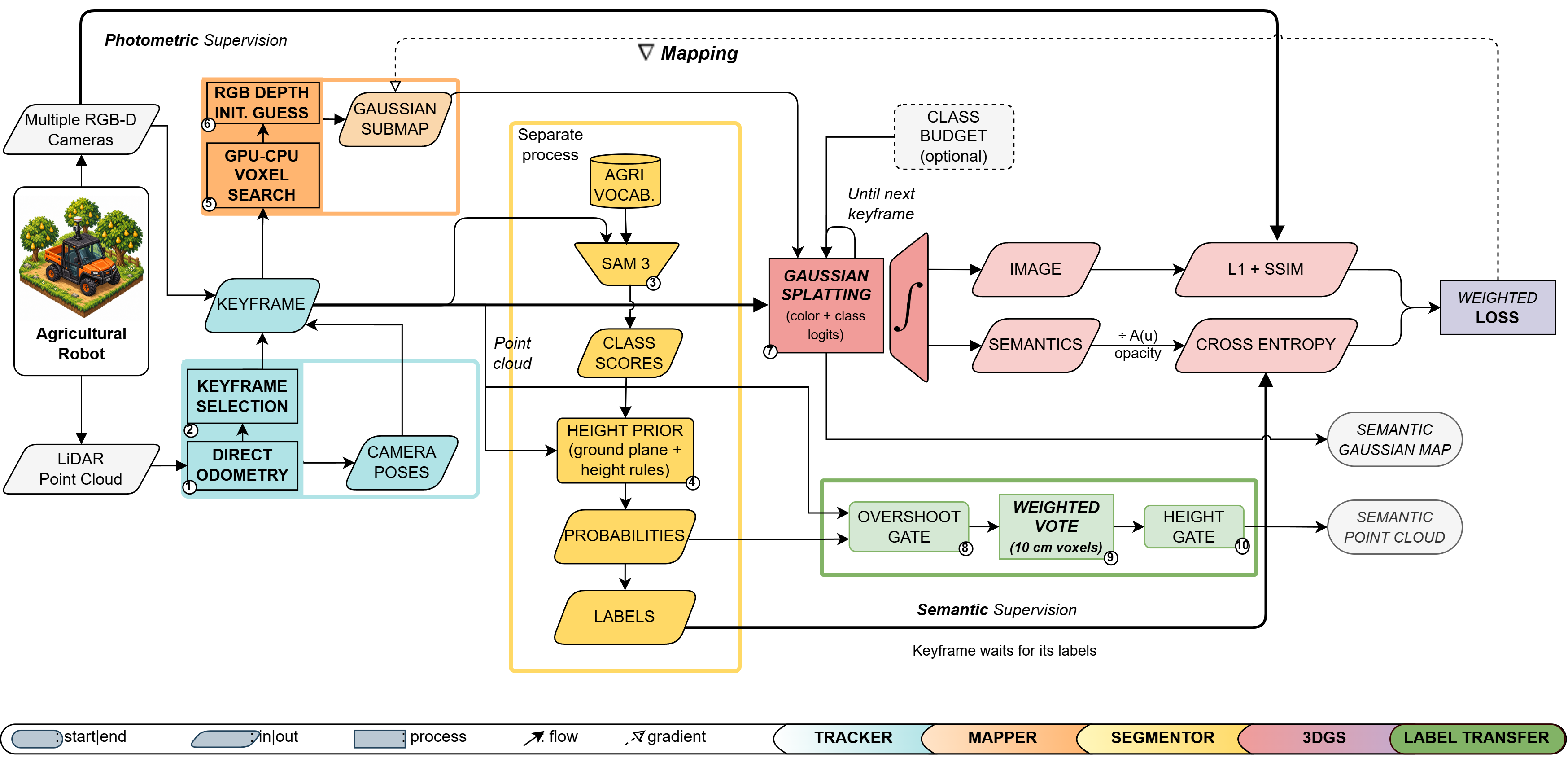}
   \caption{Overview of \ours{}. LiDAR odometry (1) tracks every scan and selects keyframes (2). In a separate process, SAM~3~\cite{carion2026sam} (3) scores the text-prompted classes on the keyframe image, and the height prior (4) vetoes classes outside their height bands above a ground plane fitted to the stereo points, yielding per-pixel class probabilities and labels. The mapper keeps only the Gaussians near the keyframe on the GPU (5) and seeds new ones from its stereo points (6). Once its labels arrive, the keyframe is optimized until the next one is selected (7): the rendered image and the opacity-normalized class map are supervised by L1+SSIM and cross-entropy, and an optional class budget (dashed) constrains refinement. A helper process lifts the class probabilities to the stereo points, discards trunk observations lying behind monocular depth (8), votes in 10\,cm voxels (9), and relabels as ground any voxel within 0.15\,m of the local ground (10), updating the semantic point cloud after every keyframe.}
   \label{fig:overview}
\end{figure*}

3D Gaussian Splatting (3DGS)~\cite{kerbl2023gaussian} has enabled photorealistic maps to be built incrementally and efficiently within modern SLAM systems~\cite{matsuki2024gaussianslam,keetha2024splatam}, including recent Visual--LiDAR Gaussian SLAM approaches deployed in real orchards~\cite{usuelli2026agrigs}. Existing orchard Gaussian maps, however, remain primarily optimized for geometric and photometric reconstruction and lack the explicit semantic information required for higher-level downstream robotic reasoning.

Transferring image semantics to 3D points is unreliable for narrow structures: Peters et al.~\cite{peters2023labeltransfer} report mIoU dropping from $66.1$ to $48.1$ after multi-view transfer, with narrow classes such as poles among those that degrade. We address this by fusing multi-view evidence and gating labels with the local ground surface and monocular depth.

Thin structures may also be systematically disadvantaged during map optimization: densification allocates new primitives according to reconstruction gradients~\cite{kerbl2023gaussian}, which, under a finite primitive budget, can favor large or highly textured regions. We therefore explicitly reserve refinement capacity for underrepresented semantic classes before distributing the remaining budget according to reconstruction demand.

We call the resulting system \ours{}, from the Latin \textit{arbor} (tree); its main contributions are as follows.

\begin{itemize}
\item To our knowledge, the first \textit{online semantic Gaussian SLAM system for orchards}, which jointly optimizes appearance and semantics within the Gaussian map without post-run refinement, extending semantic Gaussian SLAM beyond indoor RGB-D settings~\cite{sgsslam,zhu2025semgauss,gs3lam}.

\item An \textit{online semantic point cloud}, delivered alongside the Gaussian map, that fuses multi-view evidence and gates labels with the local ground surface and monocular depth, without 3D annotation, and is more accurate than naive transfer on all six sequences.

\item \textit{Class-constrained allocation of the Gaussian refinement budget}, reserving new primitives for important but spatially limited structures, which improves training-view tree accuracy only under reduced budgets.

\item A class-resolved evaluation on training views and held-out frames across two species and three phenological stages, against state-of-the-art AgriGS-SLAM and three semantic Gaussian SLAM systems.
\end{itemize}

\section{RELATED WORK}

\textbf{Gaussian Splatting SLAM.} Dense SLAM first adopted 3DGS~\cite{kerbl2023gaussian} for indoor scenes observed by monocular or RGB-D cameras~\cite{keetha2024splatam,matsuki2024gaussianslam,yugay2024gaussianslam}. AgriGS-SLAM~\cite{usuelli2026agrigs} later extended Gaussian mapping to challenging orchard environments using LiDAR odometry and multi-camera reconstruction, yet its resulting map still encodes only geometry and appearance.

\textbf{Semantic Gaussian mapping and capacity allocation.} Beyond geometry and appearance, maps increasingly carry a semantic layer derived from images. Following Semantic-NeRF~\cite{zhi2021semanticnerf}, Gaussian representations have learned semantic and instance features~\cite{zhou2024feature3dgs,ye2024gaussiangrouping}, and SGS-SLAM~\cite{sgsslam}, GS3LAM~\cite{gs3lam}, and SemGauss-SLAM~\cite{zhu2025semgauss} optimize semantics online, but in indoor RGB-D scenes where dense depth ties each pixel to a surface. Semantics now also steer map capacity in task-conditioned~\cite{gausslite} and horticultural~\cite{octosplat} mapping, yet only by separating targets from background, not by reserving capacity for several classes at once.

\textbf{Semantic fusion for thin structures.} In agriculture, semantic mapping has addressed fruit monitoring~\cite{lobefaro2025spatiotemporal} and tree mapping~\cite{cheng2023treescope}, and orchard SLAM models trees as landmarks~\cite{treeslam}. Semantic point clouds are typically built by projecting image predictions onto 3D points, aggregating them across views~\cite{peters2023labeltransfer}, and discarding points the camera cannot see~\cite{berrio2020semanticfusion}; this assumes that each pixel and its 3D point share a surface. Thin trunks and trellis posts violate this assumption: their pixels can take depth from the surface behind them, yielding wrong labels that visibility filtering cannot detect and multi-view agreement may reinforce.

\section{METHOD}
\ours{} addresses these gaps in a single online pipeline. As shown in Fig.~\ref{fig:overview}, camera poses are provided by LiDAR odometry based on scan-to-submap registration~\cite{chen2022dlo}, with keyframes selected by translation and rotation thresholds (steps 1--2); the Gaussian map does not feed back into tracking, so semantic processing cannot affect the trajectory.

The Gaussian map $\mathcal{G}=\{g_i\}_{i=1}^{N}$ represents each primitive as $g_i = (\boldsymbol{\mu}_i, \boldsymbol{\Sigma}_i, \sigma_i, \mathbf{c}_i, \mathbf{s}_i)$, with position, covariance, opacity, appearance, and a view-independent semantic logit vector $\mathbf{s}_i\in\mathbb{R}^{C}$ over $C=6$ classes: sky, ground, canopy, trunk, trellis, and fruit. Sky is assigned to pixels with no value in the precomputed monocular depth maps~\cite{depthanything}, and pixels that no class prompt explains receive an ignore index. Segmentation runs in a separate process, and each keyframe is optimized once its prediction is available, with a semantic target that depends only on its own image, stereo points, monocular depth, and the ground planes of earlier keyframes. Past keyframes are never optimized again.

\subsection{Geometry-Constrained Semantic Supervision}
\label{sec:supervision}
Open-vocabulary segmentation can serve every species and phenological stage without retraining, but it predicts from the image alone and does not explicitly account for orchard geometry. We use a frozen promptable model~\cite{carion2026sam} (step 3 in Fig.~\ref{fig:overview}) with classes supplied as text prompts: each class is scored by the maximum, over its prompt's detections, of detection score times mask probability, and the top-scoring class is selected without thresholding. We therefore constrain these scores with a simple height prior (step 4) at two levels: pixels and connected components.

\textbf{Pixel level.} For each keyframe, we fit a ground plane by RANSAC to its stereo point cloud, accepting only plausible tilt and camera height and refitting or falling back to the running median of earlier keyframes when the camera height deviates from it. Pixel heights above this plane must respect class-specific bands, a minimum for canopy and a crown ceiling for trunk: where stereo depth is valid and in range, a class outside its band has its score set to $-\infty$, so the pixel takes its best remaining class instead of being discarded.

\textbf{Component level.} The segmenter can read bare trunks as trellis posts, and no per-pixel height band can reliably separate the two, since both start at the ground; posts, however, typically reach much higher. A connected trellis component therefore keeps its predicted class only if its tallest valid pixel reaches a minimum height, and components without valid in-range depth abstain. Sky is assigned last, overriding every other label, and the segmenter sees the full image before sky masking, preserving thin backlit structures.

\subsection{Semantic Gaussian Mapping}
\label{sec:semantic_gaussian_representation}
Memory management (steps 5--6 in Fig.~\ref{fig:overview}) follows AgriGS-SLAM~\cite{usuelli2026agrigs}, with fixed 0.1\,m voxels: only Gaussians sharing a voxel with the current keyframe's stereo points stay \emph{active} on the GPU, the others being kept \emph{dormant} on the CPU. Learning semantics on this map (step 7) raises two issues, addressed here: rendered logits fade on thin structures, and learned loss weights downweight the semantic term.

\textbf{Semantic field.} As in semantic Gaussian representations ~\cite{zhou2024feature3dgs,ye2024gaussiangrouping} and SLAM~\cite{zhu2025semgauss}, every Gaussian carries a semantic state; unlike a distilled feature, $\mathbf{s}_i$ costs only $C$ floats per primitive and yields a class field by rendering, without a decoder. The class field is rendered by a second rasterization pass over the same positions, covariances, and opacities, without spherical harmonics. For pixel $u$, let $(i_1,\ldots,i_{K_u})$ denote the Gaussians covering $u$ front-to-back and $G_i(u)$ the projected 2D Gaussian. Defining the compositing weight
$w_{i_k}(u)=\sigma_{i_k}G_{i_k}(u)
\prod_{\ell<k}\left(1-\sigma_{i_\ell}G_{i_\ell}(u)\right)$,
the semantic field and accumulated opacity are
\begin{equation}
\mathbf{S}(u)=
\sum_{k=1}^{K_u}w_{i_k}(u)\mathbf{s}_{i_k},
\quad
A(u)=\sum_{k=1}^{K_u}w_{i_k}(u).
\label{eq:semantic_rendering}
\end{equation}
Low accumulated opacity attenuates the logits toward zero and their softmax toward uniform, particularly on thin or incomplete structures such as trunks and trellis. We therefore normalize the composited logits before the softmax,
\begin{equation}
\hat{\mathbf{p}}(u)
    =
    \operatorname{softmax}
    \left(
        \frac{\mathbf{S}(u)}
        {\max\left(A(u),\,\epsilon\right)}
    \right),
\qquad \epsilon=10^{-3},
\label{eq:semantic_probability}
\end{equation}
which leaves fully opaque pixels unchanged. With $y(u)$ the segmenter's prediction after the geometric constraints of Sec.~\ref{sec:supervision} and $\Omega$ the pixels without the ignore index, the semantic loss is
\begin{equation}
\mathcal{L}_{\mathrm{sem}}
=
-\frac{1}{|\Omega|}
\sum_{u\in\Omega}
w_{y(u)}\log\hat{p}_{y(u)}(u),
\label{eq:semantic_loss}
\end{equation}
where the class weights $w_c$ are uniform in all reported runs.

\textbf{Loss balancing.} The photometric term of~\cite{usuelli2026agrigs}, $\mathcal{L}_{\mathrm{c}}=0.8\,L_1+0.2\,(1-\mathrm{SSIM})$, and $\mathcal{L}_{\mathrm{sem}}$ are combined by homoscedastic uncertainty weighting~\cite{kendall2018multitask},
\begin{equation}
\mathcal{L} = \sum_{k}\left[\tfrac{1}{2}e^{-v_k}\,\mathcal{L}_k + \tfrac{1}{2}v_k\right],
\label{eq:homo}
\end{equation}
where each $v_k$ is a learnable scalar log-variance initialized at zero. Left free, \eqref{eq:homo} downweights whichever term carries the larger residual, and a semantic field supervised only through an alpha-composited render never competes with the photometric term on that basis. We therefore \emph{pin} the semantic log-variance at $v_{\mathrm{sem}}=-2.5$, a weight of $\tfrac{1}{2}e^{2.5}\approx6.1$, while the photometric one stays learnable. The logit vector $\mathbf{s}_i$ is updated only through $\mathcal{L}_{\mathrm{sem}}$ while its Gaussian is active and visible from the optimized keyframe; the fusion algorithm described later in Sec.~\ref{sec:transfer} never modifies it.

\subsection{Semantics-Aware Map Refinement}
\label{sec:semantic_map_construction}
An incremental mapper introduces a finite number of primitives at each refinement, and conventional densification~\cite{kerbl2023gaussian,usuelli2026agrigs} ranks candidates by reconstruction gradients, so large, textured regions such as ground and canopy can dominate the budget while thin but important classes receive too little capacity. We therefore condition refinement on class (dashed box in Fig.~\ref{fig:overview}), taking a candidate's class not from its logits, whose alpha-composited supervision makes their argmax too diffuse, but from its projection in the current keyframe's corrected label map, available because only visible primitives accumulate densification gradients.

\textbf{Class-budgeted allocation.} Let $B$ be the refinement budget and $\mathcal{D}_c$ the candidates (demand) of class $c$, with optional cap $\kappa_c\in[0,1]$ and floor share $\phi_c\geq0$. First, each class is bounded by $a_c=\min(|\mathcal{D}_c|,\lfloor \kappa_c B \rfloor)$, or $a_c=|\mathcal{D}_c|$ without a cap, retaining the highest-gradient candidates within capped classes. Second, floors reserve
\begin{equation}
q_c = \min\!\big(\lfloor B\,\bar{\phi}_c \rfloor,\ a_c\big),
\qquad
\bar{\phi}_c = \frac{\phi_c}{\sum_{c'}\phi_{c'}}.
\label{eq:floor}
\end{equation}
Finally, the remaining $B-\sum_c q_c$ slots are filled by global gradient rank within the bounds $a_c$, and unused reservations return to this pool; a cap may leave the refinement below $B$ if the other classes lack demand. Since~\eqref{eq:floor} depends only on the ratios between $\phi_c$, the floor policy is scale-invariant and independent of $B$. Under reduced budgets, floors increase training-view accuracy on tree classes, whereas at the full budget they reduce it (Sec.~\ref{sec:res_ablation}), so the main configuration, which uses the full budget, allocates without constraints.

\subsection{3D Label Transfer}
\label{sec:transfer}

Alongside the Gaussian map, a helper process builds the semantic point cloud (label transfer in Fig.~\ref{fig:overview}). Naive transfer labels a stereo point $\mathbf{x}$ by its first observation, $c(\mathbf{x})=\arg\max_c p_i(c\,|\,\mathbf{x})$, with $p_i$ the height-constrained class posterior at its pixel in keyframe $i$; this is unreliable on thin structures, whose pixels may take depth from the surface behind them. After every keyframe, the helper thus relabels the cloud in three steps: an overshoot gate first discards trunk observations lying behind monocular depth (step 8), a vote then fuses the remaining ones per voxel (step 9), and a height gate finally relabels near-ground voxels as ground (step 10), fixing errors that views agree on.

\textbf{Overshoot gate.} In each view, monocular depth~\cite{depthanything} is first mapped to the stereo depths by a trimmed least-squares affine fit, and trunk observations more than $t=0.5$\,m behind the estimated surface are discarded; $t$ is the value most frequently selected by two-fold temporal cross-validation on the reference frames of Sec.~\ref{sec:setup} in earlier experimental runs.

\textbf{Evidence-weighted multi-view voting.} Points in the same cell of a fixed $10$\,cm voxel grid are treated as observations of the same surface. For a voxel $\mathbf{x}$ with observations $\mathcal{V}(\mathbf{x})$,
\begin{equation}
c^{\star}(\mathbf{x}) =
\arg\max_c\;
\pi_c^{-\beta}
\sum_{i\in\mathcal{V}(\mathbf{x})}
w_i\,p_i(c\,|\,\mathbf{x}),
\quad \beta=\tfrac{1}{2},
\label{eq:vote}
\end{equation}
where $w_i$ increases with segmentation confidence and decreases with range beyond $3$\,m, and $\pi_c$ is class $c$'s share of accumulated evidence, so that $\pi_c^{-\beta}$ partially offsets how often dominant classes are observed, since $\beta=1$ can overemphasize rare classes.

\textbf{Height gate.} Each voxel's height $h(\mathbf{x})$ is measured above the local ground, the median height of the ground-labeled voxels in its 1\,m$^{2}$ horizontal cell:
\begin{equation} c(\mathbf{x}) \leftarrow \textsc{gnd} \ \text{if}\ c(\mathbf{x}) \neq \textsc{gnd},\ h(\mathbf{x}) < \tau_h = 0.15\,\mathrm{m}. \label{eq:gate} \end{equation}
The threshold separates trunk labels projected onto the ground behind the tree row from those on the row itself; cells with fewer than ten ground-labeled voxels remain unchanged.

\section{EXPERIMENTAL SETUP}
\label{sec:setup}

\subsection{Data and protocol}

We use the field dataset of~\cite{usuelli2026agrigs}, whose six sequences cover apple and pear orchards at dormancy, flowering, and harvesting, each with a training and a validation traversal and ground-truth poses. All experiments use only its inline camera; adding the platform's two side cameras did not improve semantics in any test. The monocular depth maps~\cite{depthanything} come precomputed with the dataset, and our timings do not include computing them. Sky pixels are black in the photometric target, and their LiDAR returns are discarded.

\textbf{Semantic evaluation reference.} Semantics are evaluated against an offline reference produced by the same segmenter~\cite{carion2026sam} and height bands as our system, but with prompts and post-processing tuned per species on separate manually annotated frames. Held-out scores use 157 inline-camera reference frames of the validation traversals (17 to 36 per sequence), at least 1\,s and 0.5\,m apart, disjoint from the tuning portion of each row, and rendered at their odometry poses without registration to the ground truth; training views use a reference produced identically for every training keyframe. Except for the segments of Sec.~\ref{sec:baselines}, results come from full-length runs.

\subsection{Metrics}
\label{sec:retention}

\textbf{Rendering.} We report PSNR, SSIM, and LPIPS (AlexNet) over the full image, averaged over training keyframes right after their last optimization step (\emph{training view}) and over validation keyframes (5 to 20 per sequence) rendered from the completed map without prior optimization (\emph{novel view}). Since sky is black in the photometric target, rendered content there is penalized.

\textbf{Semantics.} Pixel labels are the argmax of the rendered semantic channels (for the baselines, their released decoding), with pixels of opacity below 0.1 assigned to sky. Per-class IoU comes from a confusion matrix pooled over all evaluated pixels of a sequence, excluding reference pixels of structures outside the orchard. We report mIoU over all six classes and \emph{structural} mIoU over ground, canopy, trunk, and trellis, leaving out classes absent from both reference and prediction. The semantic cloud is scored at the same frames by projecting voxel centers with a z-buffer onto the covered pixels.

\textbf{Trajectory.} Absolute trajectory error (ATE) is the RMSE of estimated positions against the ground truth after SE(3) Umeyama alignment~\cite{umeyama1991least} without scale correction, over robot positions at keyframes for our system and AgriGS-SLAM and over inline-camera positions at every frame for the semantic baselines; training and validation traversals are aligned separately, and odometry stalls remain in the error.

\textbf{Variability.} Three repeats of the main configuration on pear harvesting on the same machine give a per-metric noise floor (e.g., 36\% of the mean for held-out trunk IoU versus 1.6\% for ground); ablation changes beyond three standard deviations are treated as effects, and trajectories are deterministic across repeats.

\subsection{Baselines and ablations}
\label{sec:baselines}

\begin{table*}[t]
\centering
\caption{Comparison with semantic Gaussian SLAM on a 301-frame segment of each sequence (inline camera), every system estimating its own poses: mIoU/structural mIoU on training views (TV) and held-out frames (HO), training ATE, and wall-clock time. All methods receive the same images, stereo depth, and segmenter labels. Bold: best; --: no completed run.}
\label{tab:h2h}
\scriptsize
\setlength{\tabcolsep}{2.2pt}
\resizebox{\textwidth}{!}{%
\begin{tabular}{l cc cc cc cc cc cc}
\toprule
 & \multicolumn{6}{c}{Pear} & \multicolumn{6}{c}{Apple} \\
\cmidrule(lr){2-7}\cmidrule(lr){8-13}
 & \multicolumn{2}{c}{Dormancy} & \multicolumn{2}{c}{Flowering} & \multicolumn{2}{c}{Harvesting} & \multicolumn{2}{c}{Dormancy} & \multicolumn{2}{c}{Flowering} & \multicolumn{2}{c}{Harvesting} \\
\cmidrule(lr){2-3}\cmidrule(lr){4-5}\cmidrule(lr){6-7}\cmidrule(lr){8-9}\cmidrule(lr){10-11}\cmidrule(lr){12-13}
Method & TV & HO & TV & HO & TV & HO & TV & HO & TV & HO & TV & HO \\
\midrule
\multicolumn{13}{l}{\textit{mIoU/structural mIoU}\,$\uparrow$} \\
SemGauss-SLAM~\cite{zhu2025semgauss} & \multicolumn{2}{c}{out of memory, frame 256} & \multicolumn{2}{c}{out of memory, frame 224} & \multicolumn{2}{c}{out of memory, frame 256} & \multicolumn{2}{c}{out of memory, frame 256} & \multicolumn{2}{c}{out of memory, frame 224} & \multicolumn{2}{c}{out of memory, frame 192} \\
SGS-SLAM~\cite{sgsslam} & .334/.291 & .056/.021 & .372/.364 & .080/.023 & .438/.457 & .144/.125 & \multicolumn{2}{c}{out of memory, frame 258} & .318/.327 & .115/.110 & .394/.431 & .182/.193 \\
GS3LAM~\cite{gs3lam} & .247/.193 & .090/.009 & .314/.288 & .114/.050 & .323/.329 & .206/.196 & .356/.298 & .166/.123 & .293/.307 & .109/.063 & .319/.340 & .187/.210 \\
\ours{} & \textbf{.742/.687} & \textbf{.407/.328} & \textbf{.775/.729} & \textbf{.435/.371} & \textbf{.665/.734} & \textbf{.421/.457} & \textbf{.756/.730} & \textbf{.409/.368} & \textbf{.748/.712} & \textbf{.397/.371} & \textbf{.630/.657} & \textbf{.339/.373} \\
\midrule
\multicolumn{13}{l}{\textit{ATE [m]}\,$\downarrow$ / \textit{time [min]}\,$\downarrow$} \\
SemGauss-SLAM & \multicolumn{2}{c}{--} & \multicolumn{2}{c}{--} & \multicolumn{2}{c}{--} & \multicolumn{2}{c}{--} & \multicolumn{2}{c}{--} & \multicolumn{2}{c}{--} \\
SGS-SLAM & \multicolumn{2}{c}{7.99 / 112} & \multicolumn{2}{c}{17.42 / 150} & \multicolumn{2}{c}{3.48 / 130} & \multicolumn{2}{c}{--} & \multicolumn{2}{c}{5.28 / 133} & \multicolumn{2}{c}{5.00 / 129} \\
GS3LAM & \multicolumn{2}{c}{22.15 / 58} & \multicolumn{2}{c}{8.09 / 64} & \multicolumn{2}{c}{3.06 / 48} & \multicolumn{2}{c}{5.17 / 39} & \multicolumn{2}{c}{8.31 / 57} & \multicolumn{2}{c}{3.72 / 54} \\
\ours{} & \multicolumn{2}{c}{\textbf{0.06} / \textbf{16}} & \multicolumn{2}{c}{\textbf{0.07} / \textbf{20}} & \multicolumn{2}{c}{\textbf{0.06} / \textbf{19}} & \multicolumn{2}{c}{\textbf{0.12} / \textbf{22}} & \multicolumn{2}{c}{\textbf{0.06} / \textbf{23}} & \multicolumn{2}{c}{\textbf{0.14} / \textbf{26}} \\
\bottomrule
\end{tabular}}
\end{table*}

\textbf{Photometric orchard SLAM.} We run AgriGS-SLAM~\cite{usuelli2026agrigs} (\textsc{AgriGS}) on all six sequences from its released code with two changes: adaptive keyframing is disabled, so both systems insert a keyframe every 1.0\,m or 15$^\circ$, and a logging fault that otherwise skips keyframes is guarded. It keeps its budget of 50k new Gaussians per refinement (ours: 75k) and its arrival-driven scheduler: like ours, it optimizes each keyframe until odometry provides the next, so step counts depend on hardware and workload (medians of 10 to 15 per keyframe, ours 9 to 15). Its maps are scored with our protocol, since its own evaluation masks the sky; apart from the trajectory errors reported in~\cite{usuelli2026agrigs}, all AgriGS results are our re-runs. Unlike the baseline, our DLO registers every LiDAR scan rather than only camera-synchronized ones and uses a second-order motion prior. For Photo-SLAM~\cite{photoslam}, Splat-SLAM~\cite{splatslam}, PINGS~\cite{pings}, OpenGS-SLAM~\cite{opengsslam}, and DLO~\cite{chen2022dlo} with 3DGS~\cite{kerbl2023gaussian}, we use the trajectory errors and, for the first four, the PSNR that~\cite{usuelli2026agrigs} reports on the same six sequences.

\textbf{Semantic Gaussian SLAM.} We also run SGS-SLAM~\cite{sgsslam}, GS3LAM~\cite{gs3lam}, and SemGauss-SLAM~\cite{zhu2025semgauss} from their released code and configurations, changing only data loading, the GPU build, and the export of training-view renders. All three receive the inline images at $960\times600$, stereo depth, and per-frame labels from our segmenter and prompts, but not our height prior, fusion, or gates, and each estimates its own poses. Because their per-frame optimization makes full routes costly, all methods process the same 301 consecutive frames of each sequence (40.4 to 48.6\,m of path), placed where the most held-out reference frames lie within 0.75\,m and 30$^\circ$ of a segment camera. Held-out scores use those reference frames, and training-view scores use the keyframes of our segment runs, rendered by every method right after it optimizes that frame. Where our truncated run drifts on the validation traversal (apple segments and pear flowering), held-out frames are rendered at the poses of our full-route run, registered to the segment map on shared training frames without ground truth. 95\% confidence intervals come from 10,000 paired bootstrap resamples over the 14 to 26 held-out frames and 39 to 47 training views of each segment.

\textbf{Ablations.} On pear harvesting, each variant of Table~\ref{tab:ablation} changes one component: \emph{no height prior}; \emph{no opacity normalization}~\eqref{eq:semantic_probability}; \emph{learned $v_{\mathrm{sem}}$} instead of the pinned value; and \emph{no semantic loss}, which leaves segmentation running. \textsc{Ours-NS} (no semantics), run on all six sequences, disables segmentation, the semantic loss, and the semantic point cloud. Class-budgeted refinement is evaluated at $B=75$k with a 15\% ground cap, trunk and trellis floors~\eqref{eq:floor}, or both (one run each), and at $B=6$k and $12$k with a 2\% sky cap, with and without floor shares of 0.17, 0.35, 0.25, 0.17, and 0.06 for ground, canopy, trunk, trellis, and fruit (three paired runs each).

\subsection{Implementation details}
The segmenter~\cite{carion2026sam} runs in a separate process on the same GPU with a $1008\times1008$ input and one text prompt per class for both species and all stages (``grassy orchard ground'', ``tree branches and leaves'', ``tree trunk'', ``vertical pole and wire'', and ``round fruit''); its class scores supervise the map at $240\times150$. Ground planes fitted to the stereo points allow at most 25$^\circ$ of tilt and 0.7 to 2.0\,m of camera height. The bands are canopy $\geq0.7$\,m and trunk $\leq1.0$\,m for pear and canopy $\geq0.6$\,m and trunk $\leq0.9$\,m for apple, the latter derived from row geometry, and trellis components must reach 0.8\,m. Each step adds up to $B=75$k Gaussians while fewer than $3.6\times10^{6}$ are active. The helper fuses every keyframe, including validation keyframes, as soon as it is uploaded. All systems run on NVIDIA GeForce RTX 4090 GPUs and AMD EPYC 7763 CPUs; variants are only compared with runs on the same machine. Code will be released upon acceptance.

\section{RESULTS}
\label{sec:results}
We compare \ours{} with semantic Gaussian SLAM (Sec.~\ref{sec:res_h2h}), evaluate it over full routes (Sec.~\ref{sec:res_main}), and ablate its components (Sec.~\ref{sec:res_ablation}).

\begin{table*}[t]
\centering
\caption{Full-route results (inline camera): ATE, PSNR of training keyframes after their last step and of validation keyframes, semantic accuracy on training views (TV) and held-out frames (HO), and structural mIoU of the semantic point cloud with naive transfer and our fusion. \textsc{AgriGS}: AgriGS-SLAM~\cite{usuelli2026agrigs} scored with our protocol; \textsc{Ours-NS}: \ours{} without semantics. Bold: lower ATE; more accurate point cloud. $^\dagger$\ours{}: mean of three runs.}
\label{tab:main}
\resizebox{\textwidth}{!}{%
\begin{tabular}{l cc cc ccc ccc ccc cc}
\toprule
 & \multicolumn{2}{c}{ATE train [m]\,$\downarrow$} & \multicolumn{2}{c}{ATE val [m]\,$\downarrow$} & \multicolumn{3}{c}{PSNR train [dB]\,$\uparrow$} & \multicolumn{3}{c}{PSNR val [dB]\,$\uparrow$} & \multicolumn{3}{c}{Semantics\,$\uparrow$} & \multicolumn{2}{c}{Point cloud\,$\uparrow$} \\
\cmidrule(lr){2-3}\cmidrule(lr){4-5}\cmidrule(lr){6-8}\cmidrule(lr){9-11}\cmidrule(lr){12-14}\cmidrule(lr){15-16}
Sequence & \textsc{AgriGS} & Ours & \textsc{AgriGS} & Ours & \textsc{AgriGS} & \textsc{Ours-NS} & Ours & \textsc{AgriGS} & \textsc{Ours-NS} & Ours & TV mIoU & HO mIoU & HO struct. & naive & fused \\
\midrule
Apple dormancy & 0.392 & \textbf{0.389} & \textbf{0.306} & 0.481 & 21.60 & 21.34 & 19.91 & 14.60 & 14.32 & 13.70 & 0.698 & 0.341 & 0.311 & 0.183 & \textbf{0.195} \\
Apple flowering & 4.404 & \textbf{0.264} & 0.706 & \textbf{0.163} & 23.09 & 22.09 & 20.63 & 14.78 & 14.17 & 13.50 & 0.680 & 0.345 & 0.329 & 0.181 & \textbf{0.192} \\
Apple harvesting & 7.983 & \textbf{0.444} & \textbf{0.306} & 0.344 & 22.94 & 22.74 & 21.29 & 15.99 & 15.65 & 14.61 & 0.614 & 0.338 & 0.396 & 0.167 & \textbf{0.174} \\
\midrule
Pear dormancy & 0.252 & \textbf{0.082} & 0.258 & \textbf{0.068} & 22.47 & 23.13 & 21.60 & 16.30 & 17.08 & 16.76 & 0.709 & 0.346 & 0.269 & 0.140 & \textbf{0.149} \\
Pear flowering & 0.280 & \textbf{0.088} & 0.290 & \textbf{0.055} & 22.96 & 23.76 & 22.50 & 18.74 & 18.79 & 18.72 & 0.758 & 0.356 & 0.299 & 0.174 & \textbf{0.185} \\
Pear harvesting$^\dagger$ & 0.332 & \textbf{0.161} & 0.229 & \textbf{0.111} & 23.21 & 22.73 & 21.54 & 16.61 & 16.31 & 15.27 & 0.655 & 0.317 & 0.346 & 0.185 & \textbf{0.196} \\
\midrule
Mean & 2.274 & \textbf{0.238} & 0.349 & \textbf{0.204} & 22.71 & 22.63 & 21.25 & 16.17 & 16.05 & 15.43 & 0.686 & 0.340 & 0.325 & 0.172 & \textbf{0.182} \\
\bottomrule
\end{tabular}}
\end{table*}

\subsection{Comparison with Semantic Gaussian SLAM}
\label{sec:res_h2h}

\begin{figure*}[!t]
   \centering
   \includegraphics[width=\textwidth]{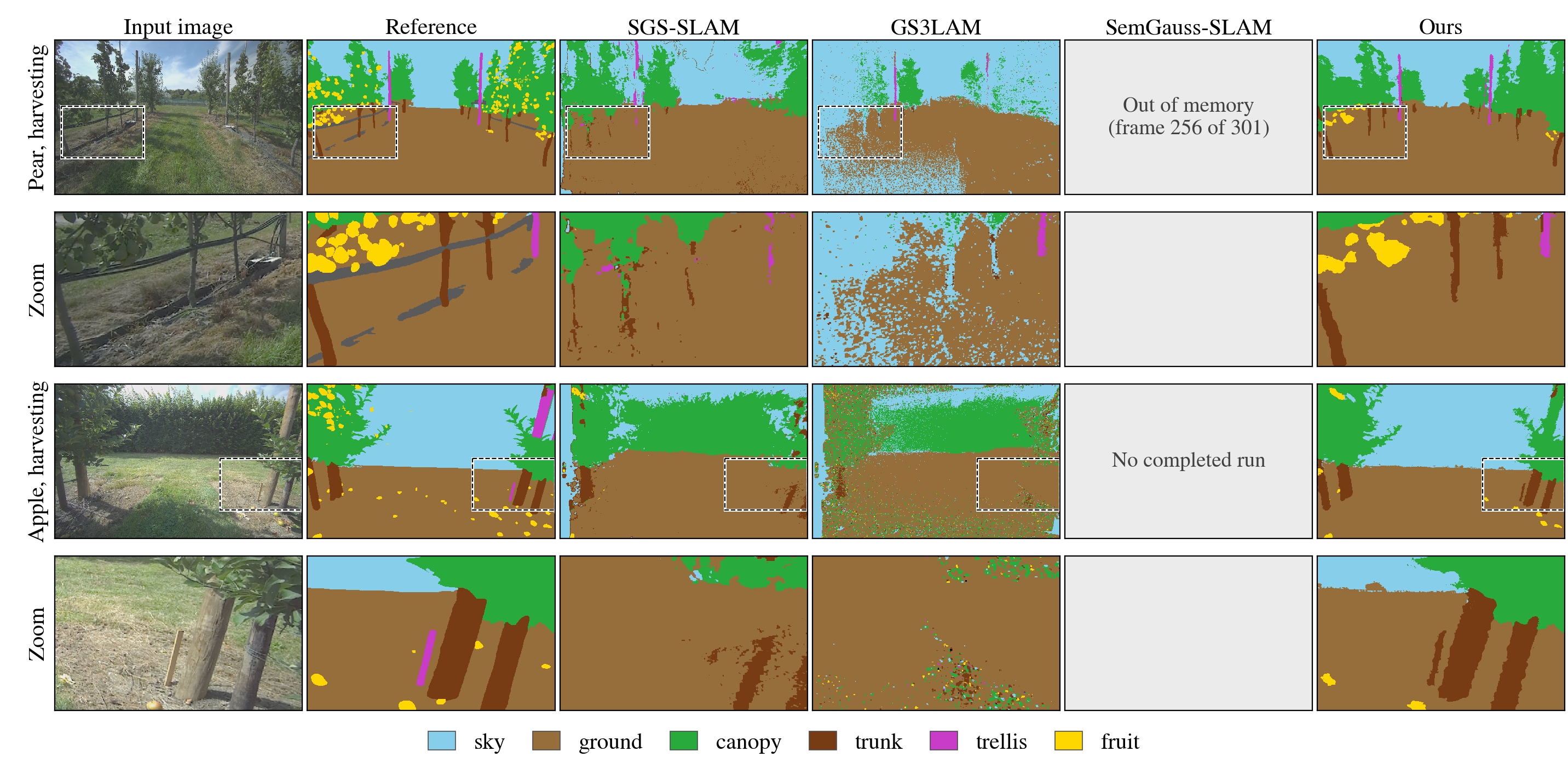}
   \caption{Selected training views of the harvesting segments of Table~\ref{tab:h2h}, rendered right after each method optimizes the frame, with zooms of the dashed boxes. The baselines lose most thin structures, which \ours{} largely recovers.}
   \label{fig:h2h_qual}
\end{figure*}

\begin{figure*}[!t]
   \centering
   \includegraphics[width=0.97\textwidth]{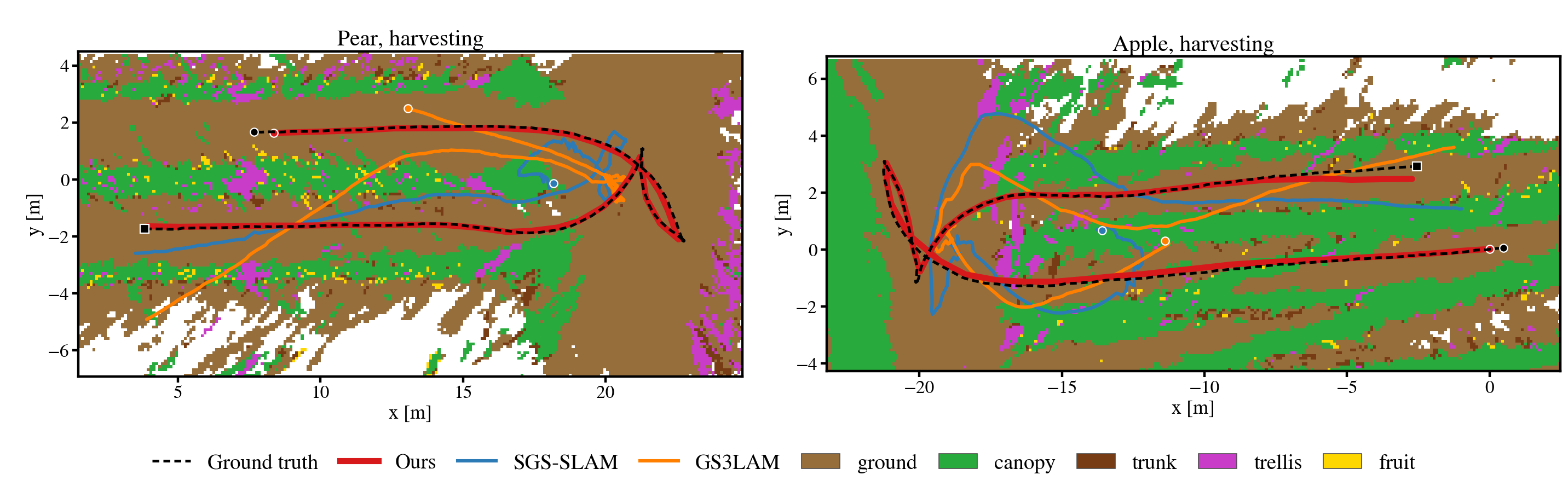}
   \caption{Trajectories of the runs in Fig.~\ref{fig:h2h_qual} after SE(3) Umeyama alignment to the ground truth, over the semantic point cloud built online by \ours{} (highest voxel per 0.1\,m cell), whose canopy rows and trellis posts outline the orchard. Squares: start; dots: end of each trajectory.}
   \label{fig:traj}
\end{figure*}

\textbf{Held-out views.} The baselines struggle with orchard-scale mapping (Table~\ref{tab:h2h}). SemGauss-SLAM exhausts the 24\,GB GPU in bundle adjustment on all six segments, and SGS-SLAM during mapping on one. Where GS3LAM and SGS-SLAM complete, their RGB-D tracking, designed for indoor scenes, drifts by 3.06 to 22.15\,m, against 0.06 to 0.14\,m for our LiDAR odometry. In Fig.~\ref{fig:traj}, the robot drives down one alley, turns at the row end, and returns along the next: \ours{} retraces this path, including the turn, whereas both baselines cut through tree rows, and SGS-SLAM ends inside a row in both segments. \ours{} therefore leads on the held-out frames of every completed segment by 0.15 to 0.36 mIoU and 0.16 to 0.35 structural mIoU, with all 95\% intervals excluding zero. Thin structures account for much of this margin: on the held-out frames of the pear harvesting segment, \ours{} reaches trunk and trellis IoU of 0.12 and 0.28, while neither baseline exceeds 0.01.

\textbf{Training views.} Drift alone does not explain this gap: on views rendered right after their optimization, where tracking errors matter least, \ours{} surpasses the baselines on every completed segment by $+0.23$ to $+0.50$ mIoU and $+0.23$ to $+0.49$ structural mIoU, with all 95\% intervals excluding zero, and reaches trunk IoU of 0.44 to 0.62 against at most 0.12 for any baseline, whose maps lose most thin structures. In Fig.~\ref{fig:h2h_qual}, \ours{} keeps each pear trunk as a continuous line from canopy to ground and recovers both apple trunks in the zoom, where SGS-SLAM leaves only fragments and the render of GS3LAM breaks into speckles; it still misses the thin trellis stake beside them.

\textbf{Efficiency.} This accuracy does not cost time: \ours{} is also the fastest system, 1.7 to 3.7 times faster than GS3LAM and 5.0 to 7.5 times faster than SGS-SLAM where they complete (Table~\ref{tab:h2h}), although only its time includes segmentation, the validation traversal, and the semantic point cloud.

\begin{table}[t]
\centering
\caption{Single-camera ATE per sequence and mean PSNR on training (TV) and novel views (NV). $^\dagger$Reported in~\cite{usuelli2026agrigs}, whose PSNR excludes sky error; $^*$partial trajectory.}
\label{tab:sota}
\scriptsize
\setlength{\tabcolsep}{2.2pt}
\resizebox{\columnwidth}{!}{%
\begin{tabular}{l ccc ccc cc}
\toprule
 & \multicolumn{6}{c}{ATE [m]\,$\downarrow$} & \multicolumn{2}{c}{PSNR [dB]\,$\uparrow$} \\
\cmidrule(lr){2-7}\cmidrule(lr){8-9}
 & \multicolumn{3}{c}{Apple} & \multicolumn{3}{c}{Pear} & & \\
\cmidrule(lr){2-4}\cmidrule(lr){5-7}
Method & Dorm. & Flow. & Harv. & Dorm. & Flow. & Harv. & TV & NV \\
\midrule
Photo-SLAM$^\dagger$~\cite{photoslam} & 18.95 & 15.16 & 19.10 & 16.24 & 23.72 & 19.00 & 8.42 & 8.67 \\
Splat-SLAM$^\dagger$~\cite{splatslam} & 5.26 & 28.14 & 21.89 & 6.64$^*$ & 20.85 & 18.29 & 20.45 & 12.78 \\
PINGS$^\dagger$~\cite{pings} & 171.50 & 132.94 & 89.44 & 157.23 & 87.35 & 97.76 & 9.39 & 9.07 \\
OpenGS-SLAM$^\dagger$~\cite{opengsslam} & 20.70 & 20.95 & 20.22 & 21.07 & 10.93 & 20.82 & 14.04 & 14.93 \\
DLO+3DGS$^\dagger$ & 0.576 & 0.408 & 0.658 & 0.440 & 0.448 & 0.755 & & \\
AgriGS-SLAM$^\dagger$ & 0.519 & 0.354 & 0.737 & 0.426 & 0.440 & 0.720 & & \\
\midrule
\ours{}, training & \textbf{0.389} & 0.264 & 0.444 & 0.082 & 0.088 & 0.161 & \textbf{21.25} & \\
\ours{}, validation & 0.481 & \textbf{0.163} & \textbf{0.344} & \textbf{0.068} & \textbf{0.055} & \textbf{0.111} & & \textbf{15.43} \\
\bottomrule
\end{tabular}}
\end{table}

\subsection{Full-Route Mapping across Species and Seasons}
\label{sec:res_main}

\begin{figure*}[!t]
   \centering
   \includegraphics[width=0.8\textwidth]{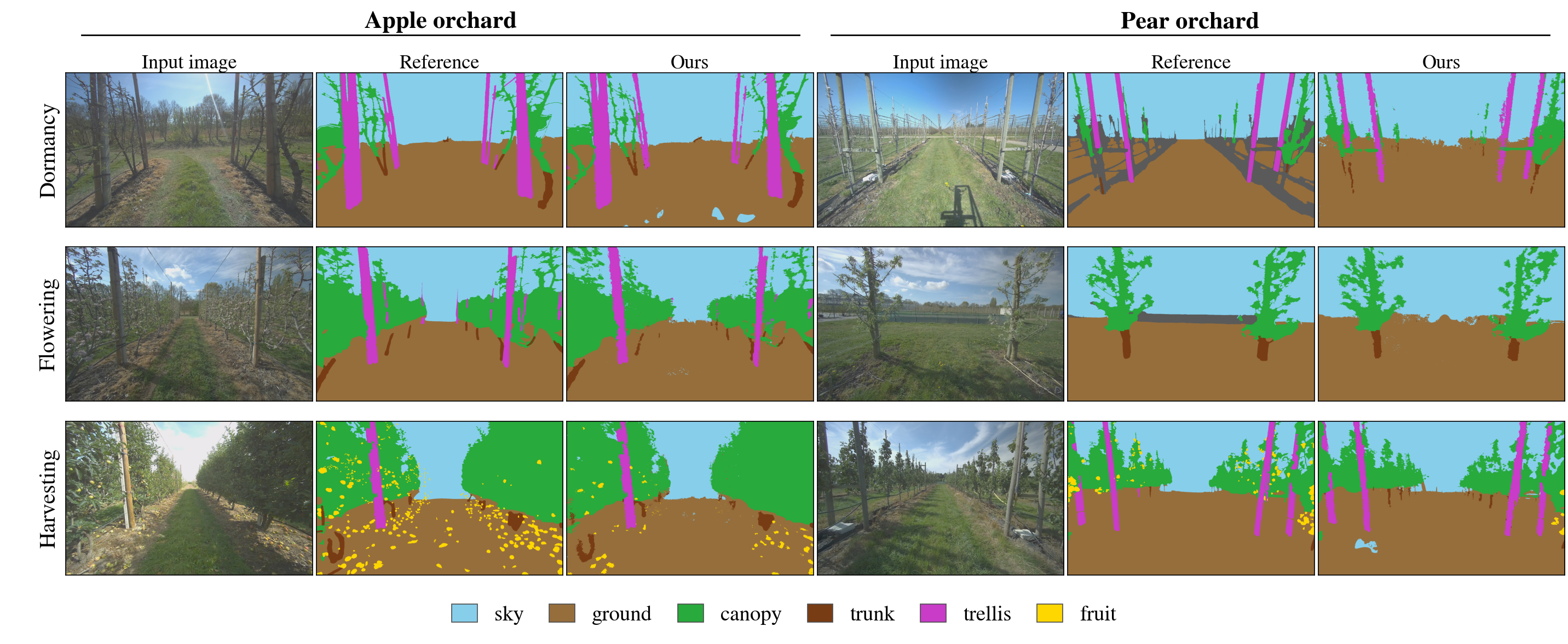}
   \caption{Selected training views of \ours{} on full routes across species and stages, with one set of text prompts.}
   \label{fig:seasons}
\end{figure*}

\textbf{Tracking.} Over full routes, LiDAR odometry keeps \ours{} below 0.5\,m ATE on all 12 traversals, and disabling semantics leaves all trajectories unchanged. \ours{} achieves lower ATE than AgriGS-SLAM on all six training traversals of Table~\ref{tab:main} and on four of six validation traversals, and AgriGS-SLAM loses track on apple harvesting in our replay. The general-purpose 3DGS-SLAM systems evaluated in~\cite{usuelli2026agrigs} reach 5.3 to 171.5\,m (Table~\ref{tab:sota}), and our training ATE is lower than the values reported for DLO+3DGS and AgriGS-SLAM on all six sequences.

\textbf{Rendering.} Semantics costs some image quality, but \ours{} still renders better than the general-purpose systems of~\cite{usuelli2026agrigs}. Without semantics, our mapper matches AgriGS-SLAM in mean PSNR at comparable step budgets (Table~\ref{tab:main}) but trails it in LPIPS (0.44 against 0.41 on training and 0.57 against 0.54 on validation keyframes) and training SSIM (0.64 against 0.69). Learning semantics costs 1.39\,dB of training PSNR on average and 0.63\,dB on validation keyframes, on pear harvesting mostly through the semantic loss itself (Sec.~\ref{sec:res_ablation}). Those systems report lower PSNR than ours on 44 of their 48 per-sequence values (Table~\ref{tab:sota}), although their evaluation excludes sky error.

\textbf{Semantics.} One set of text prompts serves both species and all stages, but novel views remain much harder than training views: held-out mIoU stays between 0.317 and 0.356, while training views reach 0.614 to 0.758. Across the stages of Fig.~\ref{fig:seasons}, from bare branches and posts in dormancy to fruit on canopy and ground at harvest, trellis posts stay intact, while isolated fruit and thin branch tips are where the renders lose most detail. Label fusion helps consistently but modestly: the online semantic point cloud is more accurate than naive first-view transfer on every sequence, with structural mIoU gains of $+0.007$ to $+0.013$ on the 4.5 to 5.8\% of reference pixels covered by projected voxels.

\begin{table}[t]
\centering
\caption{Ablations on pear harvesting. First row: \ours{}, mean and standard deviation (sd) of three runs; other rows: change from that mean (one run each). TV: training views; HO: held-out frames; struct.: structural mIoU. Bold: change beyond three sd.}
\label{tab:ablation}
\scriptsize
\setlength{\tabcolsep}{3pt}
\begin{tabular}{l cc ccc c}
\toprule
 & \multicolumn{2}{c}{TV\,$\uparrow$} & \multicolumn{3}{c}{HO\,$\uparrow$} & PSNR\,$\uparrow$ \\
\cmidrule(lr){2-3}\cmidrule(lr){4-6}
Variant & struct. & trunk & struct. & canopy & trellis & train [dB] \\
\midrule
\ours{} & 0.733 & 0.539 & 0.346 & 0.429 & 0.121 & 21.54 \\
\quad sd & 0.002 & 0.004 & 0.007 & 0.004 & 0.010 & 0.06 \\
\midrule
No height prior & \textbf{$-$0.040} & \textbf{$-$0.103} & $-$0.015 & \textbf{$-$0.041} & $+$0.002 & $+$0.10 \\
No opacity norm. & $-$0.005 & $-$0.005 & \textbf{$-$0.040} & \textbf{$-$0.046} & \textbf{$-$0.043} & \textbf{$+$0.21} \\
Learned $v_{\mathrm{sem}}$ & \textbf{$-$0.033} & \textbf{$-$0.104} & 0.000 & \textbf{$+$0.015} & $-$0.010 & \textbf{$+$0.94} \\
No semantic loss & \multicolumn{5}{c}{no semantic field is learned} & \textbf{$+$0.91} \\
\textsc{Ours-NS} & \multicolumn{5}{c}{no semantics} & \textbf{$+$1.19} \\
\bottomrule
\end{tabular}
\end{table}

\subsection{Ablation Study}
\label{sec:res_ablation}

Each component of \ours{} protects a different part of the map (Table~\ref{tab:ablation}). \textbf{Opacity normalization} matters most on held-out views: removing it lowers structural mIoU by 0.040, while training views barely change. \textbf{The height prior} is essential for canopy and trunk: without it, held-out canopy IoU drops by 0.041 and training-view trunk IoU by 0.103. \textbf{Pinning the semantic weight} protects thin structures: when learned, the weight decays to 0.28, about one sixth of the color weight, and the 0.94\,dB of PSNR gained is paid for with 0.104 of training-view trunk IoU. \textbf{The semantic loss} causes most of the photometric cost of semantics: disabling it recovers 0.91 of the 1.19\,dB separating \ours{} from \textsc{Ours-NS}.

\textbf{Class-budgeted refinement} allocates primitives as specified but helps only under reduced budgets. With floors at $B=12$k, the shares of new Gaussians granted to trunk and trellis rise by 0.17 and 0.10 in all three runs. Since floors take capacity from ground and sky, which dominate the image, we assess them by \emph{tree mIoU}, over canopy, trunk, trellis, and fruit. Under reduced budgets, floors raise training-view tree mIoU in two of three paired runs at both 6k (mean $+0.05$) and 12k (mean $+0.03$), and fruit IoU in all six pairs (mean $+0.05$), while mean held-out tree mIoU changes by $+0.001$ and $-0.006$. At the full budget of 75k, floors lower training-view tree mIoU by 0.024, so \ours{} uses unconstrained allocation.

\textbf{Cost of semantics.} Semantics adds no measurable wall-clock time: on the same machines, \ours{} and \textsc{Ours-NS} differ in run time by $-7.2\%$ to $+2.6\%$ (mean $-0.9\%$), as the mapper optimizes each keyframe until odometry releases the next. The cost appears instead in optimization: with semantics, the mapper completes 8 to 28\% fewer steps in the same time, and the final maps are 1.16 to 1.57 times smaller, holding 68 to 95 million Gaussians. In an instrumented pear harvesting run, keyframes start every 5.96\,s, segmentation is available after 0.56\,s, and the helper fuses a keyframe in 250\,ms without blocking the mapper (medians).

\section{CONCLUSIONS}
Orchard robots need maps that are semantic as well as spatial, in every season. We presented \ours{}, an online semantic Gaussian Splatting SLAM system that builds such maps with one set of text prompts and without post-run refinement, and our experiments support each of its contributions. First, among the semantic Gaussian SLAM systems evaluated, only \ours{} maps orchard rows reliably, exceeding SGS-SLAM and GS3LAM on every completed segment by 0.15 to 0.36 held-out and 0.23 to 0.50 training-view mIoU while running 1.7 to 7.5 times faster. Second, its semantic point cloud is more accurate than naive transfer on all six sequences. Third, under reduced budgets, class-budgeted refinement raises training-view fruit IoU in all six paired runs and tree mIoU in four. Finally, our class-resolved evaluation across two species and three stages locates the remaining gap in thin structures: on full routes, held-out trunk IoU stays below 0.07 on every sequence, against 0.41 to 0.56 on training views. Future work will target their novel-view semantics by revisiting past keyframes and closing loops, so that robots returning to a row can rely on the map from new viewpoints.





\bibliographystyle{IEEEtran}
\bibliography{IEEEabrv, mybib}

\end{document}